\documentclass[11pt]{article}
\usepackage[margin=1in]{geometry}
\usepackage{amsmath,amssymb}
\usepackage{graphicx}
\usepackage{float}
\usepackage{booktabs}
\usepackage{hyperref}
\usepackage{natbib}
\usepackage{xcolor}

\hypersetup{colorlinks=true, linkcolor=blue, citecolor=blue, urlcolor=blue}

\title{Same Geometry, Opposite Noise:\\Transformer Magnitude Representations Lack Scalar Variability}

\author{Jon-Paul Cacioli\\
Independent Researcher, Melbourne, Australia\\
\textit{Classical Minds, Modern Machines}}

\date{}

\begin{document}
\maketitle

\begin{abstract}
Scalar variability---the finding that representational noise scales proportionally with magnitude, producing a constant coefficient of variation---is a hallmark of biological magnitude systems. We tested whether transformer language models exhibit this property by analysing the dispersion of hidden-state representations across carrier sentences for 26 numerical magnitudes in three 7--8B parameter models (Llama-3-8B-Instruct, Mistral-7B-Instruct-v0.3, Llama-3-8B-Base; data from \citealt{cacioli2026weber}). We found the opposite: representational variability \textit{decreased} with magnitude along the magnitude axis (scaling exponent $\alpha \approx -0.19$; 0/16 primary layers with $\alpha > 0$, all three models). The negative sign was consistent in full-dimensional space ($\alpha \approx -0.04$) and after sentence-identity correction ($\alpha \approx -0.007$). The anti-scalar pattern was 3--5$\times$ stronger along the magnitude axis than orthogonal dimensions, and corpus frequency strongly predicted per-magnitude variability ($\rho = .84$). These results demonstrate that distributional learning alone is insufficient to produce scalar variability: transformers reproduce log-compressive magnitude geometry but not the constant-CV noise signature observed in biological systems.


\smallskip
\noindent Pre-registered on OSF prior to analysis: \url{https://osf.io/w4892}. All results reported regardless of outcome.
\end{abstract}

\section{Introduction}

Biological magnitude systems share two properties. First, representational geometry is log-compressive: internal distances follow a logarithmic function of stimulus magnitude, producing Weber's Law \citep{ganguli2014efficient}. Second, representational noise scales proportionally with magnitude, producing a constant coefficient of variation---the scalar property \citep{gibbon1977scalar, gallistel2000nonverbal, meck1983mode}.

Under the efficient coding framework, both properties arise jointly when a system with a fixed noise budget encodes a power-law-distributed input: the distributional statistics produce the geometry, while the capacity constraint produces the noise scaling. These two properties are theoretically dissociable. A system that shares the distributional precondition but lacks the capacity constraint should exhibit the geometry without the noise.

Transformer language models provide exactly this test case. \citet{cacioli2026weber} demonstrated that three 7--8B parameter transformers encode numerical magnitude with log-compressive geometry (RSA $\rho = .68$--.96, Stevens $\beta \approx 0.01$), consistent with efficient coding under the power-law distribution of integers in training data ($\alpha = 0.77$). But transformers have no metabolic budget---no firing-rate bound, no population-size limit, no fixed noise allocation.

We pre-registered four hypotheses. H1: representational variability increases with magnitude ($\alpha > 0$). H2: the scaling exponent approximates 1 (constant CV). H3: the exponent is below 1, reflecting the absence of metabolic constraints. H4: the exponent varies with layer depth.

\section{Method}

\subsection{Data}

All data were collected for \citet{cacioli2026weber} and are publicly archived. Hidden-state vectors were extracted at the magnitude token position (pre-RMSNorm) from Llama-3-8B-Instruct, Mistral-7B-Instruct-v0.3, and Llama-3-8B-Base at all 33 layers for 26 numerical magnitudes (\{1, 2, \ldots, 9, 10, 15, 20, \ldots, 90, 100, 150, 200, 300, 500, 700, 1000\}) in 5 carrier sentences.

\subsection{Measures}

The primary variability measure was the mean Euclidean distance of the 5 carrier-sentence vectors from their centroid at each magnitude and layer: $V(n,l) = \frac{1}{5}\sum_i \|h_i(n,l) - \mu(n,l)\|_2$.

Two co-primary controls were pre-registered. First, a sentence-identity correction ($V_\text{residual}$): the mean vector per sentence across all 26 magnitudes was subtracted before computing $V$, removing systematic sentence-level shifts. Second, a magnitude-axis projection ($V_\text{proj}$): variance of the 5 vectors projected onto PC1 of the 26 centroids, isolating magnitude-relevant dispersion.

At each of 16 primary layers (transformer layers 16--31, matching the peak-geometry range in \citealt{cacioli2026weber}), $\log(V)$ was regressed on $\log(n)$ via OLS to estimate the scaling exponent $\alpha$, with Theil--Sen as a robustness check. Bootstrap 95\% CIs (10{,}000 resamples, seed = 42). Under scalar variability, $\alpha = 1$. Pre-registered outlier exclusion: $V > 3\times$ layerwise median.

Dispersion across 5 fixed carrier sentences is a proxy for representational variability, not a direct measure of stochastic noise; the sentence-identity control mitigates but does not fully eliminate systematic contextual contributions.

\subsection{Pre-Registration}

All hypotheses, measures, and exclusion rules were registered on OSF prior to analysis (\url{https://osf.io/w4892}). The author had observed centroid representations (for RSA in the parent study) but had not computed any variance or dispersion measures on the per-sentence hidden states.

\section{Results}

\subsection{Confirmatory}

\begin{figure}[H]
\centering
\includegraphics[width=\textwidth]{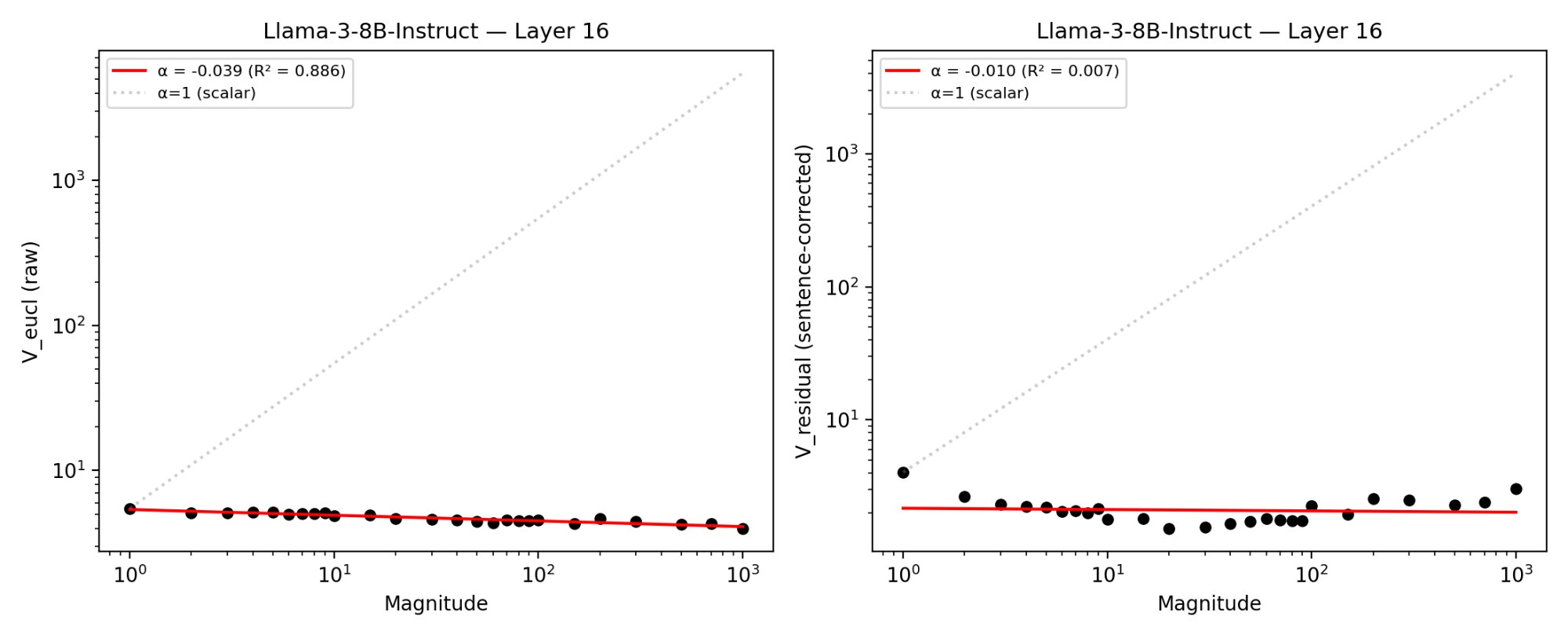}
\caption{Representational variability as a function of numerical magnitude (log--log axes) at layer 16 of Llama-3-8B-Instruct. Left: raw $V_\text{eucl}$. Right: sentence-corrected $V_\text{residual}$. Red line: OLS fit. Grey dotted line: scalar prediction ($\alpha = 1$). Data cluster near constant or slightly declining variability, far below the scalar prediction.}
\label{fig:vmag}
\end{figure}

\textbf{H1 ($\alpha > 0$) was not supported.} The scaling exponent was negative at all 16 primary layers in all three models and across all three measures (Table~\ref{tab:results}; Figure~\ref{fig:vmag}). Variability \textit{decreased} with magnitude. Mean $\alpha$ ranged from $-0.031$ (Llama-Base) to $-0.046$ (Llama-Instruct), with the raw-measure estimates implying a $\sim$10--15\% decrease in variability across the full magnitude range. No outlier cells met the pre-registered $3\times$ median criterion. OLS and Theil--Sen estimators agreed closely (mean $|\Delta\alpha| < 0.003$).

\textbf{H2 ($\alpha \approx 1$) was not supported.} No layer in any model produced $\alpha$ within $[0.8, 1.2]$.

\textbf{H3 ($\alpha < 1$) was supported} at all 48 model--layer cells. The result was stronger than predicted: $\alpha$ was not merely sub-scalar but anti-scalar.

\textbf{H4 (layerwise profile) was not supported.} Spearman $\rho$ between layer depth and $\alpha$ was non-significant in all models ($\rho = -.12$ to $.09$, all $p > .60$). The anti-scalar pattern was stable across depth.

\begin{table}[H]
\centering
\caption{Scaling exponent $\alpha$ (mean across 16 primary layers). $V_\text{eucl}$: mean Euclidean distance from centroid. $V_\text{residual}$: same, after removing sentence-level means. $V_\text{proj}$: SD along magnitude axis (PC1).}
\label{tab:results}
\begin{tabular}{lccc}
\toprule
\textbf{Model} & $\boldsymbol{V_\textbf{eucl}}$ & $\boldsymbol{V_\textbf{residual}}$ & $\boldsymbol{V_\textbf{proj}}$ \\
\midrule
Llama-3-8B-Instruct & $-0.046$ & $-0.009$ & $-0.189$ \\
Mistral-7B-Instruct & $-0.042$ & $-0.007$ & $-0.210$ \\
Llama-3-8B-Base     & $-0.031$ & $-0.005$ & $-0.085$ \\
\bottomrule
\end{tabular}
\end{table}

\subsection{Exploratory}

\begin{figure}[H]
\centering
\includegraphics[width=0.6\textwidth]{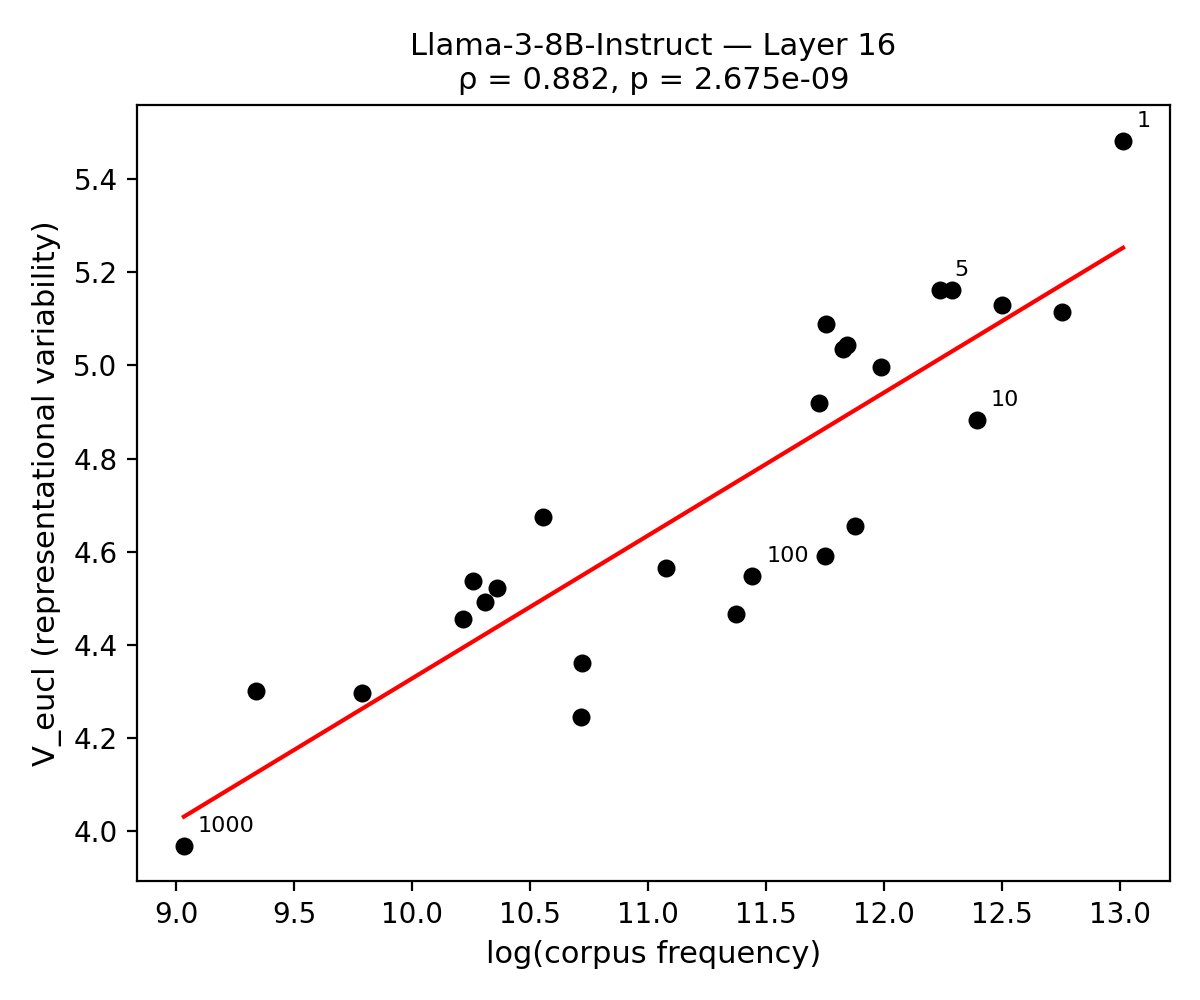}
\caption{Corpus frequency predicts representational variability. Each point is one of 26 numerical magnitudes at layer 16 of Llama-3-8B-Instruct. Numbers appearing more frequently in training data show greater representational dispersion. Spearman $\rho = .882$, $p < 10^{-9}$. Labels indicate selected magnitudes.}
\label{fig:freq}
\end{figure}

\textbf{Corpus frequency (E5).} Per-magnitude variability was strongly correlated with log corpus frequency (Spearman $\rho = .83$--.85, $p < .001$ at all 16 primary layers in all three models; Figure~\ref{fig:freq}). This pattern is consistent with a distributional account: frequent numbers (predominantly small) appear in more diverse linguistic contexts, producing wider representational dispersion; rare numbers (predominantly large) appear in narrower contexts, producing tighter clusters.

\begin{figure}[H]
\centering
\includegraphics[width=0.7\textwidth]{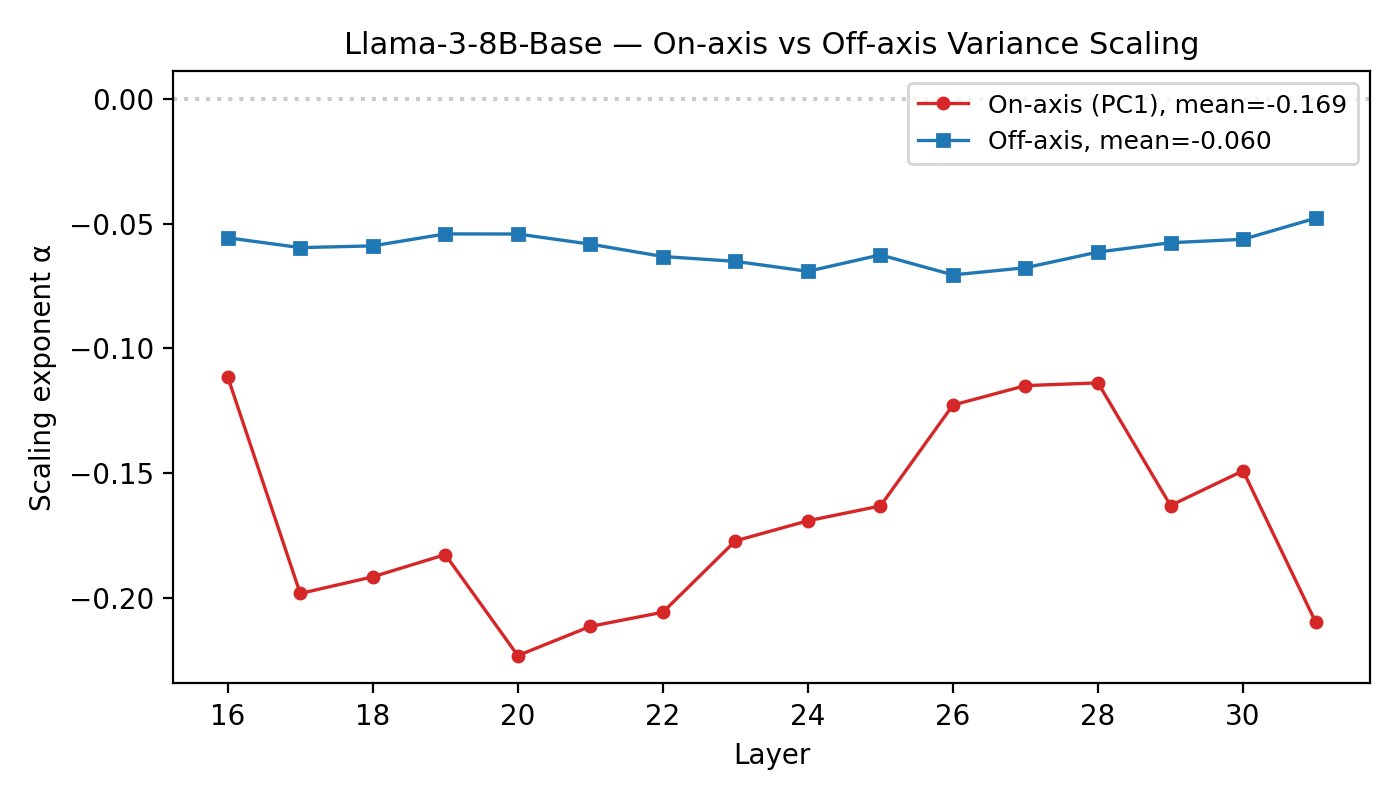}
\caption{E4: On-axis (PC1, magnitude direction) vs off-axis scaling exponent across primary layers for Llama-3-8B-Base. The anti-scalar pattern is $\sim$3$\times$ stronger along the magnitude axis (mean $\alpha = -0.169$) than orthogonal dimensions (mean $\alpha = -0.060$). Wilcoxon $p < .001$. All models showed the same dissociation.}
\label{fig:e4}
\end{figure}

\textbf{Magnitude-axis decomposition (E4).} The anti-scalar pattern was 3--5$\times$ stronger along the magnitude axis (PC1) than orthogonal dimensions (on-axis $\alpha = -0.17$ to $-0.42$; off-axis $\alpha = -0.06$ to $-0.09$; Wilcoxon $p < .001$ for all models; Figure~\ref{fig:e4}). The variability reduction at large magnitudes is concentrated in the behaviourally relevant subspace.

\begin{figure}[H]
\centering
\includegraphics[width=\textwidth]{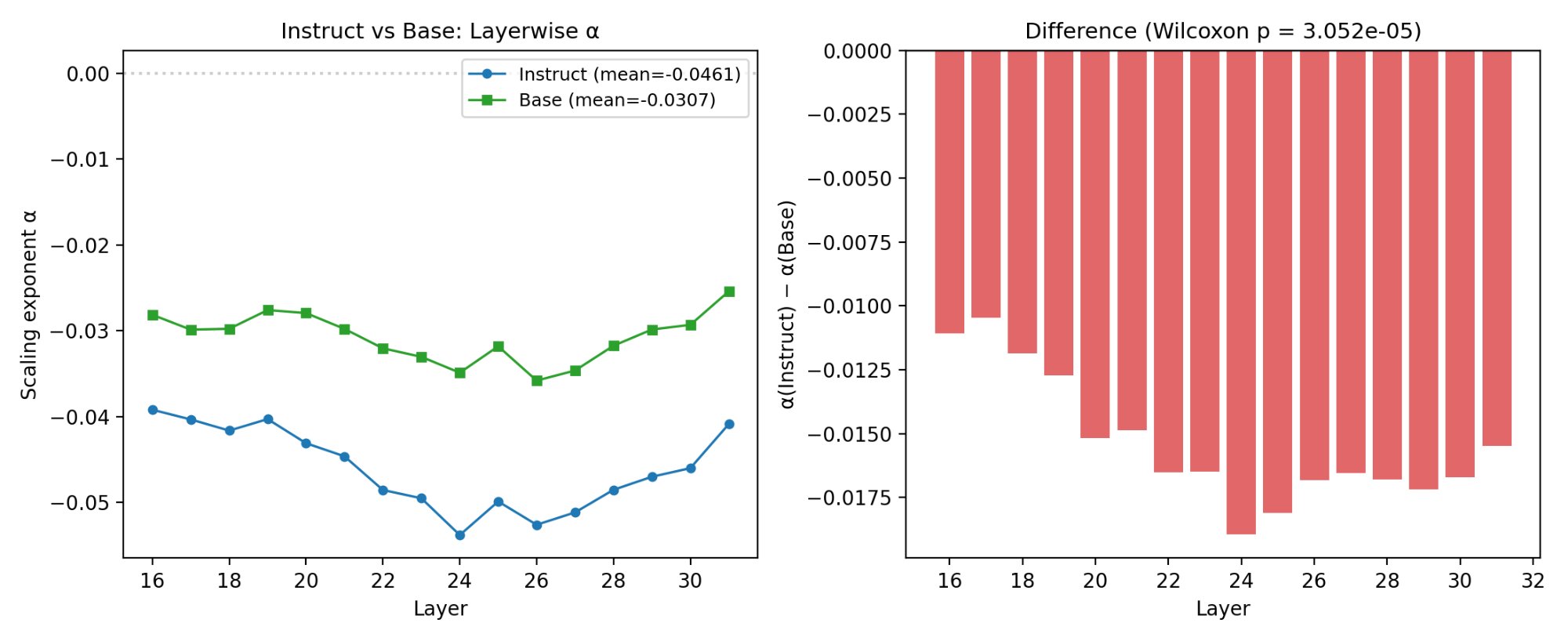}
\caption{E6: Instruction tuning amplifies the anti-scalar pattern. Left: layerwise $\alpha$ for Llama-Instruct (mean $= -0.046$) vs Llama-Base (mean $= -0.031$). Right: per-layer difference ($\Delta\alpha < 0$ at all 16 layers, Wilcoxon $p < .001$).}
\label{fig:e6}
\end{figure}

\textbf{Instruction tuning (E6).} Llama-Instruct showed a more negative $\alpha$ than Llama-Base ($\Delta\alpha = -0.015$, Wilcoxon $p < .001$ across 16 layers; Figure~\ref{fig:e6}). Instruction tuning amplifies the anti-scalar pattern, further tightening large-number representations.

\begin{figure}[H]
\centering
\includegraphics[width=\textwidth]{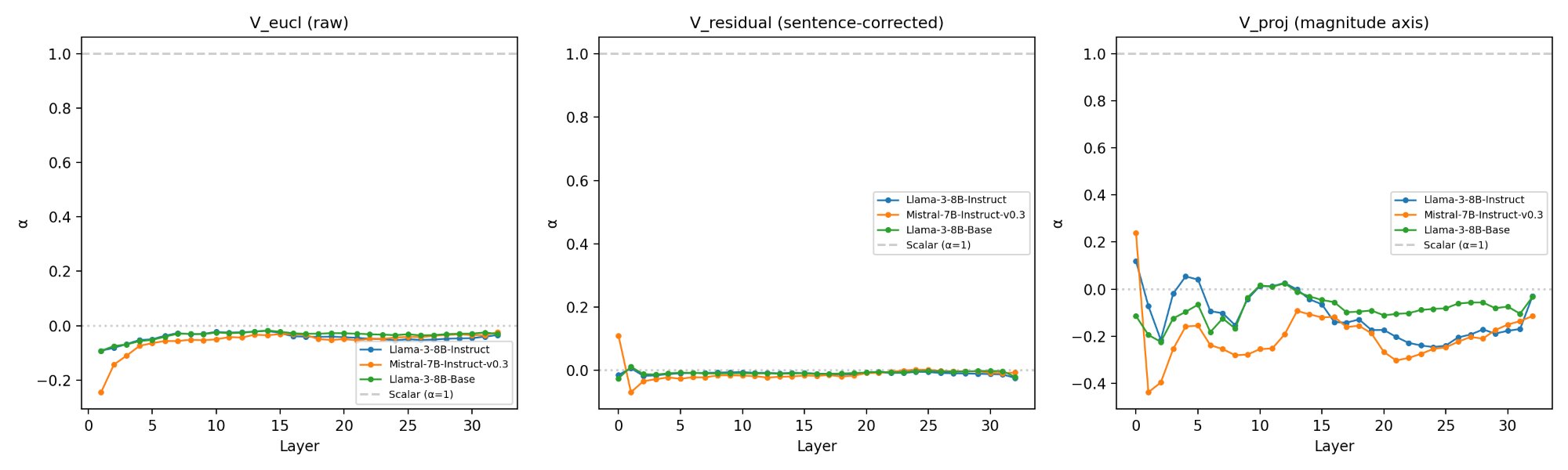}
\caption{Scaling exponent $\alpha$ across all layers for all three models. Left: $V_\text{eucl}$ (raw). Centre: $V_\text{residual}$ (sentence-corrected). Right: $V_\text{proj}$ (magnitude axis). Dashed grey line: scalar prediction ($\alpha = 1$). Dotted grey line: $\alpha = 0$. All models show $\alpha < 0$ at all primary layers across all measures.}
\label{fig:alpha}
\end{figure}

\textbf{Layerwise profiles (all models).} Figure~\ref{fig:alpha} shows the scaling exponent across all 33 layers for all three models and all three variability measures. The anti-scalar pattern is present at all layers (not just primary), is consistent across models, and is strongest on the magnitude-axis projection ($V_\text{proj}$, right panel).

\section{Discussion}

Transformer language models reproduce the mean geometry of biological magnitude systems---log-compressive encoding consistent with efficient coding \citep{cacioli2026weber}---but not the noise structure. Where biological systems exhibit scalar variability (constant CV; \citealt{gibbon1977scalar}), transformers exhibit the opposite: representational variability decreases with magnitude.

This dissociation suggests that the two properties attributed to efficient coding are separately determined. The distributional statistics of the input (power-law frequency of integers) are sufficient to produce log-compressive geometry under gradient-based optimisation. But they are insufficient to produce scalar noise. In biological systems, scalar variability arises from the interaction of efficient coding with metabolic constraints---fixed noise budgets that force proportional allocation \citep{pardo2019mechanistic}. Without such constraints, transformers allocate less variability to large magnitudes, driven by the distributional structure of training contexts rather than a capacity limitation.

The corpus frequency correlation ($\rho \approx .84$) is consistent with this account: variability tracks contextual diversity, not magnitude per se. Small numbers appear in thousands of distinct constructions; large numbers appear in narrower, more stereotyped contexts. The magnitude-axis decomposition (E4) shows that this effect is concentrated in the behaviourally relevant subspace, and the instruction-tuning comparison (E6) shows that fine-tuning further compresses rare-number representations.

Dispersion across 5 carrier sentences is a proxy for representational variability, not a direct measure of stochastic noise, and per-magnitude variance estimates from $n = 5$ are imprecise. However, the sign of $\alpha$ was negative at all 48 model--layer cells (3 models $\times$ 16 layers), all three variability measures, and under both OLS and Theil--Sen estimators. Under a null of $\alpha = 0$ with independent sign, this consistency has a binomial probability below $10^{-14}$---the conclusion does not depend on the precision of individual estimates. Generalisation to larger models and non-numerical domains is untested.

In sum, distributional learning produces the geometry of biological magnitude representation but not the noise. Scalar variability may require the metabolic constraints that transformers lack.

\section*{Open Science Statement}

Pre-registered on OSF prior to analysis: \url{https://osf.io/w4892}. All data archived with \citet{cacioli2026weber}. Analysis code: \url{https://github.com/synthiumjp/weber}. All results reported regardless of outcome.

\bibliographystyle{apalike}
\bibliography{references}

\end{document}